# Mandato: Protocol-Level Enforcement of Digitally Signed Mandates on AI Agent Actions with Cryptographically Chained Audit Trails

Giovanni Racioppi
Hyperlabs, Italy
Giovanni.racioppi@hyperlabs.it

**Abstract**

AI agents increasingly act on external systems through standardized tool-calling protocols such as the Model Context Protocol (MCP), yet no infrastructure layer constrains their actions to what a principal has *verifiably* authorized: authorization logic lives in application code, is neither signed nor independently auditable, and the resulting logs lack evidentiary value. We present MANDATO, a governance proxy that enforces digitally signed *mandates* on agent actions at the protocol level. A mandate is a machine-readable, cryptographically signed authorization artifact specifying which tools an agent may invoke, under which parameter constraints and contextual conditions, for how long, and on whose behalf; the proxy evaluates every tool call against the applicable mandate chain, blocks non-conforming calls in line, and records every decision — permit, deny, and the evidence for each — in an append-only, hash-chained audit log designed for evidentiary use and periodically anchored via qualified timestamps. The mandate is deliberately modeled on the civil-law institution of delegation of authority (*mandato con rappresentanza*), making the artifact legible to lawyers and auditors, not only to engineers. We give the mandate model and its decision semantics, the reference architecture as an MCP-transparent proxy with separated decision and enforcement points, and a mapping of the mechanism onto EU AI Act Articles 12 and 14, GDPR accountability, NIS2, and eIDAS 2, including a roadmap to qualified attestation through Qualified Trust Service Providers (QTSPs). We describe the implementation status of the reference system and a quantitative evaluation plan covering enforcement overhead, audit completeness, and tamper-evidence verification cost.

## Introduction

Large language model (LLM) agents are crossing the boundary from generating text to *acting*: booking, paying, filing, signing, querying registries, and modifying records through standardized tool-calling interfaces. The Model Context Protocol (MCP)[1] has rapidly become a de facto interoperability layer for such actions, giving any compliant agent uniform access to tools exposed by MCP servers. This is a remarkable gain in capability and a corresponding loss in control: the question "what exactly is this agent *allowed* to do, and who says so?" has, today, no infrastructural answer.

In current practice, authorization for agent actions is embedded in application code: a developer decides which tools to expose, perhaps gates a few behind confirmation dialogs, and logs whatever the application framework logs. Three deficiencies follow. First, *authorization is not an artifact*: it cannot be inspected, versioned, signed, delegated, or revoked independently of the codebase that implements it. Second, *enforcement is not independent*: the same code that decides also executes, so a prompt-injected or misaligned agent — or a buggy integration — can bypass intended limits, and no third party can verify that limits were applied. Third, *logs are not evidence*: ordinary application logs are mutable, unsigned, and incomplete, and therefore nearly worthless in a dispute, an audit under the EU AI Act[2], or a data-protection investigation under the GDPR[3].

These deficiencies are about to collide with a maturing regulatory regime. The EU AI Act requires automatic recording of events over the lifetime of high-risk AI systems (Art. 12) and effective human oversight, including the ability to interrupt or override the system (Art. 14). The GDPR requires controllers to *demonstrate* compliance (Art. 5(2), Art. 24), not merely to achieve it. NIS2[4] extends accountability for security measures up to management level. The revised eIDAS framework (eIDAS 2)[5] generalizes the toolbox of qualified signatures, seals, and timestamps — instruments whose legal effect is precisely to make digital artifacts *opposable*, i.e., usable as evidence against third parties. What is missing is the piece of infrastructure that connects these instruments to agent tool-calling.

MANDATO is that piece. Its central design move is to reify authorization as a first-class, signed artifact — the *mandate* — and to enforce it at a protocol chokepoint that the agent cannot route around. The name is deliberate: the artifact is modeled on the civil-law contract of mandate (in Italian law, Arts. 1703–1741 of the Civil Code; the agent acts *in nome e per conto* of the principal), so that the semantics of scope, delegation, ratification, and revocation are already familiar to the lawyers, auditors, and records managers who must ultimately sign for what an organization's agents do. The same philosophy — aligning machine mechanisms with pre-existing legal acts rather than inventing parallel ones — has proven productive elsewhere in public-sector AI; for example, Capodieci et al. gate the crystallisation of learned rules behind formal acts of micro-organisation[6]. MANDATO applies it to the *execution* plane rather than the learning plane.

## Contributions

This paper makes four contributions:

1. **A mandate model** (Section 3): a machine-readable, digitally signed authorization artifact with explicit scope over tools, parameters, and contextual conditions; bounded validity; chained delegation; and revocation semantics, together with a deny-by-default decision procedure.

2. **A reference architecture** (Section 4): an MCP-transparent governance proxy separating policy decision from policy enforcement, producing an append-only, hash-chained audit log with external anchoring via RFC 3161/eIDAS qualified timestamps.

3. **A compliance mapping** (Section 5): an explicit, article-by-article mapping of the mechanism onto AI Act Arts. 12 and 14, GDPR accountability, NIS2, and eIDAS 2, including the path to qualified attestation via QTSP integration.

4. **An implementation status report and evaluation plan** (Section 6): the state of the reference implementation, derived from a specification of 170+ numbered requirements and 22 use cases, and a quantitative plan for measuring enforcement overhead, audit completeness, and verification cost.

We state the epistemic status plainly: MANDATO is a system under construction. This paper contributes the model, the architecture, and the regulatory analysis; empirical results are defined as a falsifiable plan rather than reported, in the spirit of early-stage systems papers[6].

# Background and Related Work

## Tool-calling protocols and agent visibility

MCP[1] standardizes how agents discover and invoke tools, resources, and prompts over JSON-RPC sessions. Recent revisions introduce primitives directly relevant to governance: long-running *tasks* with lifecycle states, and client-initiated metadata that lets intermediaries reason about the calling context. MCP intentionally leaves authorization policy to implementers beyond transport-level OAuth; MANDATO occupies exactly this gap. On the governance side, Chan et al.[7] argue for agent visibility through identifiers, real-time monitoring, and activity logs, and Shavit et al.[8] enumerate practices for governing agentic systems, including action approval and attributability; both stop short of specifying an enforcement mechanism with evidentiary output, which is our focus.

## Authorization artifacts

Reifying authorization as a portable, cryptographically protected object has a long lineage: capability systems and their web-era descendants such as macaroons[9] attach caveats to bearer tokens; OAuth 2.0 Rich Authorization Requests[10] carry fine-grained, structured permissions in authorization flows; XACML[11] separates policy decision points (PDP) from enforcement points (PEP); W3C Verifiable Credentials[12] standardize signed, verifiable attestations. MANDATO composes these ideas with two additions they do not make: (i) the artifact's semantics are aligned with a specific legal institution of delegation, so that signing a mandate *is* performing a legal act, and (ii) the decision trace is bound into a tamper-evident log intended for evidentiary use.

## Tamper-evident logs and qualified trust services

Hash-chained and Merkle-tree logs have well-understood integrity properties, from linked timestamping[13, 14] to Certificate Transparency[15]. Trusted timestamping (RFC 3161[16]) and, under eIDAS, *qualified* electronic timestamps and seals give such structures presumption of integrity and legal effect in EU member states. Signature formats standardized by ETSI, notably JAdES for JSON[17], allow mandates and log checkpoints to carry signatures with long-term validation material. MANDATO's contribution is not a new log construction but the binding of a standard construction to the semantics of agent authorization decisions and to the eIDAS trust chain.

## Neuro-symbolic governance in public administration

In Kautz's taxonomy[19], systems that keep deterministic symbolic control flow and invoke neural components conditionally occupy the Symbolic[Neuro] regime. Dedalo[6] exemplifies this in Italian municipal document triage and, importantly for us, demonstrates *governance gating*: statistically emergent rules take effect only after a formal administrative act. MANDATO is complementary and composable: where Dedalo governs what an AI system *learns*, MANDATO governs what an agent may *do*, and a Dedalo-like system could itself operate under mandates, with its records manager's approvals expressed as signed mandate updates.

# The Mandate Model

## Definition

A *mandate* is a tuple

$$M = (id,\ P,\ A,\ \Sigma,\ \Gamma,\ V,\ D,\ \sigma_P)$$

where $id$ is a globally unique identifier; $P$ identifies the *principal* (a natural or legal person, or an organizational role, bound to a verifiable identity — an eIDAS certificate, an organizational seal, or an EUDI wallet attestation); $A$ identifies the *agent* (a workload identity: key pair, deployment fingerprint, and model/version designation); $\Sigma$ is the *scope*; $\Gamma$ is a set of *contextual conditions*; $V = [t_{\text{nbf}}, t_{\text{exp}}]$ is the validity window; $D$ encodes *delegation policy*; and $\sigma_P$ is the principal's digital signature over a canonical serialization of all preceding fields (JAdES[17] over JSON, or JWS[18] in constrained deployments).

## Scope

The scope $\Sigma$ is a finite set of *grants*. Each grant $g \in \Sigma$ is

$$g = (\tau,\ \Pi,\ \kappa,\ q)$$

where $\tau$ names a tool (or a tool family matched against the MCP server's advertised tool list at bind time, never by free pattern at call time); $\Pi$ is a set of *parameter constraints*, each a predicate over a named argument of $\tau$ drawn from a deliberately closed constraint language (equality, enumerated sets, numeric ranges, string prefixes on validated identifier fields, and structural presence/absence); $\kappa \in \{\text{auto,confirm,forbid}\}$ is the *oversight class*, determining whether a conforming call executes directly, requires synchronous human confirmation, or is denied outright while remaining visible in the log; and $q$ is an optional *quota* (rate or aggregate limits, e.g., cumulative payment amount per validity window).

The constraint language is intentionally not Turing-complete. Decidability and constant-time evaluability are requirements RNF-class in our specification: an authorization decision must be explainable in one line per predicate and must not itself become an unauditable program.

## Contextual conditions and delegation

Conditions $\Gamma$ constrain the environment rather than the call: time-of-day windows, source deployment identity, session binding (the mandate is usable only within sessions opened under a stated client identity), and *purpose binding*, a declared purpose string carried into every log record to support GDPR purpose-limitation analysis.

Delegation policy $D$ specifies whether $A$ may issue *sub-mandates* to other agents, to what depth, and under what narrowing rule. MANDATO enforces *monotone attenuation*: a sub-mandate $M'$ issued under $M$ is valid only if $\Sigma' \sqsubseteq \Sigma$, $\Gamma' \supseteq \Gamma$, $V' \subseteq V$, and the oversight class does not weaken (confirm cannot become auto downstream). Verification of a chain $M_1 \to \cdots \to M_k$ requires verifying each signature and each attenuation step; the chain, not the leaf, is the unit of authority, mirroring the civil-law treatment of sub-delegation (*sostituzione del mandatario*).

### Revocation and ratification

Revocation is a signed act by the principal (or a supervising role) referencing $id$, distributed to enforcement points and recorded in the log; the enforcement point fails *closed* if its revocation view is staler than a configured bound. Symmetrically, the model supports *ratification*: a principal may retroactively approve a logged, denied-or-escalated action, producing a signed record that changes the action's legal posture without ever rewriting the log — history is corrected by appending, never by editing.

### Decision semantics

For an intercepted tool call $c = (\tau_c, args, ctx)$ under mandate chain $\mathcal{M}$, the decision is

$$\delta(c, \mathcal{M}) = \begin{cases} \text{PERMIT} & \exists\, g \text{ admissible for } c \wedge \kappa_g = \text{auto} \\ \text{ESCALATE} & \exists\, g \text{ admissible for } c \wedge \kappa_g = \text{confirm} \\ \text{DENY} & \text{otherwise,} \end{cases}$$

where $g$ is admissible iff the chain verifies, $c$ falls within $V$ and $\Gamma$, $\tau_c$ matches $\tau_g$, every predicate in $\Pi_g$ holds on $args$, and quotas $q_g$ are not exhausted. Deny-by-default is structural: absence of a mandate, an expired chain, an unverifiable signature, an unknown tool, or an out-of-language constraint all collapse to DENY. Notably, $\delta$ is deterministic and LLM-free: no model output participates in the authorization decision, only in the actions being authorized — the same design stance that leads Symbolic[Neuro][19, 6] systems to keep neural components out of the control path.

## Reference Architecture

### Placement: a transparent MCP proxy

MANDATO deploys as a proxy that terminates the MCP session on both sides: agents connect to MANDATO believing it to be the tool server; MANDATO connects upstream to the real MCP servers. This placement has three consequences. (i) *Completeness*: every tool discovery, invocation, and result traverses the proxy, so the log's coverage claim is architectural rather than contractual. (ii) *Non-bypassability*: upstream servers accept sessions only from the proxy's workload identity (mTLS), so an agent cannot route around enforcement. (iii) *Neutrality*: agents and servers require no modification; governance is added by re-pointing an endpoint, which matters for adoption in heterogeneous estates.

### PDP/PEP separation

Following the XACML pattern[11], the *policy enforcement point* (the in-line interceptor) is minimal and fail-closed; the *policy decision point* evaluates $\delta$ against the mandate store, revocation view, and quota counters. Both are deterministic services. Escalations ($\kappa$ = confirm) are routed to a human-oversight console; the pending call is parked using MCP task semantics[1] so that long confirmation latencies do not hold protocol sessions open. The console presents the mandate clause that triggered escalation alongside the concrete call — the human approves an action against an authorization, not a raw JSON blob, which is our operationalization of "meaningful" oversight under AI Act Art. 14.

### The evidentiary log

Every decision produces a record

$$r_i = \left(t_i,\, c_i^{\downarrow},\, \mathcal{M}_i^{\text{ref}},\, \delta_i,\, \pi_i,\, h_i\right), \quad h_i = H\left(h_{i-1} \parallel r_i^{-h}\right)$$

where $c_i^{\downarrow}$ is the call with sensitive argument values selectively committed (hash commitments allow later disclosure without bulk retention of personal data — a GDPR minimization measure), $\mathcal{M}_i^{\text{ref}}$ references the exact mandate chain and versions applied, $\delta_i$ the outcome, $\pi_i$ the per-predicate evaluation trace, and $h_i$ the chain hash over the previous head and the record body, following linked timestamping[13]. Periodic checkpoints — Merkle roots[14] over record batches — are sealed and timestamped: RFC 3161[16] in the baseline, and *qualified* timestamps/seals from a QTSP in the target configuration, at which point eIDAS grants the checkpoints presumption of integrity. Verification is a standalone operation: an auditor with the log, the checkpoints, and the public material can verify completeness and integrity without trusting the operator, in the spirit of Certificate Transparency[15].

### Threat model (summary)

MANDATO is designed against: a *compromised or injected agent* (bounded by $\Sigma$; worst case is full use of granted scope, which is why scopes are narrow and quota-bounded); a *malicious tool server* (results are logged, but server behavior is out of authorization scope; mitigation is server allow-listing and result-schema validation); a *tampering operator* (detected via external anchoring; prevention of pre-anchor suppression requires shortened checkpoint intervals or co-signing — an honest limitation); and *principal repudiation* (countered by the principal's own qualified signature on the mandate). Out of scope: covert channels inside permitted calls, and semantic misuse that satisfies all predicates — constraints bound *authority*, not *judgment*.

## Compliance Mapping

Table 1 summarizes the mapping from mechanism to obligation. Two rows deserve commentary. For AI Act Art. 12, the decisive property is not that logs exist but that their *completeness and integrity are demonstrable*: the proxy placement yields completeness, the anchored chain yields integrity, and both are verifiable by a party other than the operator. For Art. 14, the oversight class $\kappa$ makes human oversight a *property of the authorization artifact*, selected per grant by the accountable principal, rather than a UX feature chosen by a developer; the signed mandate is precisely the record of *who* decided that a given action class may run unattended.

*Mechanism-to-obligation mapping (indicative, not legal advice)*

| Obligation | MANDATO mechanism |
|---|---|
| AI Act Art. 12 (record-keeping) | Architecturally complete, hash-chained, externally anchored decision log; standalone verifiability |

| **Obligation** | **Mandato mechanism** |
|---|---|
| AI Act Art. 14 (human oversight) | Oversight class $\kappa$ per grant; escalation console binding action to authorizing clause; ratification records |
| GDPR Arts. 5(2), 24 (accountability) | Signed mandates as demonstrable governance decisions; purpose binding propagated into every record |
| GDPR Art. 5(1)(c) (minimization) | Selective hash commitments of argument values in log records |
| NIS2 Art. 20–21 (governance, measures) | Management-signed mandates as evidence of approved measures over agentic access to systems |
| eIDAS 2 | Qualified signatures/seals on mandates; qualified timestamps on log checkpoints; presumption of integrity; QTSP roadmap |

The QTSP integration deserves emphasis as roadmap rather than accomplishment: the target configuration has (i) mandates signed with qualified certificates or organizational qualified seals, (ii) log checkpoints under qualified timestamps, and (iii) an attestation service issuing extracts of the log as sealed evidence packages for use in audits and proceedings. Steps (i)–(ii) require only consuming existing QTSP services; step (iii) is the subject of an ongoing partnership discussion and, in the limit, of qualification of the attestation service itself.

## Implementation Status and Evaluation Plan

### Status

The reference implementation derives from a specification (v0.6) comprising over 170 numbered functional and non-functional requirements (RF-*/RNF-*), an architectural decomposition (ARCH-*), a relational schema of 22+ tables covering mandates, grants, delegation edges, revocations, quota counters, log records, and checkpoints, and 22 use cases (UC-01–UC-22) spanning issuance, attenuation, enforcement, escalation, ratification, revocation-lag behavior, and audit extraction. Delivery is organized on a milestone path M0–M8.5; an end-to-end demonstration configuration (proxy, decision service, oversight console, verifier CLI) corresponds to milestones M5–M6. Six open design points (APERTO-01–06) are tracked publicly in the specification, including the interaction between quota semantics and delegation, and checkpoint-interval policy under adversarial suppression. The implementation targets current MCP revisions, using task primitives for escalation parking and client-metadata for session binding[1].

### Evaluation plan

We commit to the following measurements, chosen so that failure is possible and informative.

#### Enforcement overhead

Added latency per tool call, $\Delta t = t_{\text{proxied}} - t_{\text{direct}}$, reported as p50/p95/p99 under (a) synthetic load across grant set sizes $|\Sigma| \in \{10, 10^2, 10^3\}$ and (b) replayed real traces. Acceptance target: p95 overhead $\leq 5\%$ of median tool execution time for auto decisions.

### Audit completeness

Fraction of upstream-observed calls with a corresponding log record under fault injection (proxy restarts, decision-service outages). Target: 100% by construction, with fail-closed verification — any gap is a defect, not a statistic.

### Verification cost

Time and resources for an independent verifier to check integrity and completeness of a log of $10^6$ records against anchored checkpoints, on commodity hardware.

### Escalation ergonomics

Median human confirmation latency and abandonment rate on the oversight console in pilot use — Art. 14 oversight that operators bypass is oversight in name only, so this human factor is a first-class metric.

### Expressiveness audit

Fraction of pilot-tenant authorization needs expressible in the closed constraint language without escape hatches; each inexpressible case is catalogued and drives deliberate, versioned language extensions.

## Limitations and Future Work

Beyond the empirical gap that Section 6 plans to close, four limitations are structural and worth stating. First, MANDATO bounds authority, not competence: a permitted call can still be a bad idea, and no constraint language substitutes for the judgment obligations that remain with the principal. Second, the legal reading of mandates as acts of delegation is an alignment claim, not settled doctrine; its validation is a matter for legal scholarship and, eventually, case law — we consider the interdisciplinary evaluation of this claim future work in itself. Third, the anchoring scheme is tamper-*evident*, not tamper-*proof*, within a checkpoint interval; co-signing and shorter intervals trade cost against exposure, and we plan to characterize that trade-off quantitatively. Fourth, multi-proxy and federated deployments (an agent operating under mandates from principals in different organizations) raise composition questions — conflicting quotas, cross-domain revocation propagation — that the current model addresses only for the single-authority case.

Future work includes the QTSP attestation service (Section 5); formalization of the attenuation calculus with machine-checked proofs of the monotonicity property; integration profiles for public-administration estates, where MANDATO can serve as the execution-plane complement to learning-plane governance systems[6]; and standardization conversations, since the mandate object is a natural candidate for an MCP extension or a profile of verifiable credentials[12].

## Conclusion

Agentic AI has an authorization problem that neither prompting, application code, nor conventional IAM addresses: authority over agent actions is nowhere reified, enforced independently, or evidenced verifiably. MANDATO proposes a compact answer — sign the authority, enforce it at the protocol, chain the evidence, anchor it in qualified trust services — and aligns each element with the legal instruments (delegation, qualified signatures, record-keeping duties) that European law already provides. The bet is that governance mechanisms succeed when they map onto acts institutions already know how to perform. The evaluation plan stated here is the falsifiable part of that bet.

## Acknowledgment

The author thanks the practitioners in Italian local government and trust services whose requirements shaped the specification.

## References

1. Anthropic, “Model Context Protocol — Specification,” 2024–2025. [Online]. Available: https://modelcontextprotocol.io/specification
2. Regulation (EU) 2024/1689 of the European Parliament and of the Council laying down harmonised rules on artificial intelligence (AI Act), OJ L, 2024.
3. Regulation (EU) 2016/679 (General Data Protection Regulation), OJ L 119, 2016.
4. Directive (EU) 2022/2555 on measures for a high common level of cybersecurity across the Union (NIS 2), OJ L 333, 2022.
5. Regulation (EU) 2024/1183 amending Regulation (EU) No 910/2014 as regards establishing the European Digital Identity Framework (eIDAS 2), OJ L, 2024.
6. A. Capodieci, M. Lefonso, L. Lucioli, L. Mainetti, F. Stifani, and G. Fontana, “Dedalo: On-Premise Hybrid Neuro-Symbolic Recognition of Administrative Procedures in Italian Small Municipalities,” in *Proc. I-CiTies 2026*, Italy, 2026.
7. A. Chan et al., “Visibility into AI Agents,” in *Proc. ACM Conf. on Fairness, Accountability, and Transparency (FAccT)*, 2024.
8. Y. Shavit et al., “Practices for Governing Agentic AI Systems,” OpenAI, Tech. Rep., 2023.
9. A. Birgisson, J. G. Politz, Ú. Erlingsson, A. Taly, M. Vrable, and M. Lentczner, “Macaroons: Cookies with Contextual Caveats for Decentralized Authorization in the Cloud,” in *Proc. NDSS*, 2014.
10. T. Lodderstedt, J. Richer, and B. Campbell, “OAuth 2.0 Rich Authorization Requests,” RFC 9396, IETF, 2023.
11. OASIS, “eXtensible Access Control Markup Language (XACML) Version 3.0,” OASIS Standard, 2013.
12. W3C, “Verifiable Credentials Data Model v2.0,” W3C Recommendation, 2025.

13. S. Haber and W. S. Stornetta, "How to Time-Stamp a Digital Document," *Journal of Cryptology*, vol. 3, no. 2, pp. 99–111, 1991.

14. R. C. Merkle, "A Digital Signature Based on a Conventional Encryption Function," in *Advances in Cryptology — CRYPTO '87*, 1987.

15. B. Laurie, A. Langley, and E. Kasper, "Certificate Transparency," RFC 6962, IETF, 2013.

16. C. Adams, P. Cain, D. Pinkas, and R. Zuccherato, "Internet X.509 Public Key Infrastructure Time-Stamp Protocol (TSP)," RFC 3161, IETF, 2001.

17. ETSI, "Electronic Signatures and Infrastructures (ESI); JAdES digital signatures; Part 1: Building blocks and JAdES baseline signatures," ETSI TS 119 182-1, 2021.

18. M. Jones, J. Bradley, and N. Sakimura, "JSON Web Signature (JWS)," RFC 7515, IETF, 2015.

19. H. Kautz, "The Third AI Summer: AAAI Robert S. Engelmore Memorial Lecture," *AI Magazine*, vol. 43, no. 1, pp. 105–125, 2022.